\documentclass{article}
\usepackage{spconf,amsmath,graphicx,hyperref}
\usepackage{hyperref}
\usepackage{url}
\usepackage{algorithmic}
\usepackage{algorithm}
\usepackage{tikz}
\usepackage{subcaption} 
\usepackage{pgfplots}   
\usepackage{booktabs}      
\usepackage{multirow}      
\usepackage{tabularx}      
\usepackage[most]{tcolorbox}
\usepackage{markdown}
\usepackage{bm}

\pgfplotsset{compat=1.18} 

\title{Utilising Gradient-Based Proposals Within Sequential Monte Carlo Samplers for Training of Partial Bayesian Neural Networks} %
\name{Andrew Millard$^{*\ddagger}$, Joshua Murphy$^{*\ddagger}$, Zheng Zhao$^{\dagger}$, Simon Maskell$^{*}$
\thanks{$^\ddagger$These authors contributed equally. \\
AM and JM were funded by a Research Studentship jointly funded by the EPSRC Centre for Doctoral Training in Distributed Algorithms EP/S023445/1. This work was funded by Dstl in collaboration with the Royal Academy of Engineering via "Dstl-RAEng Research Chair in Information Fusion" under task RQ0000040616. SM thanks Dstl, UK MOD and the Royal Academy of Engineering for supporting this work. The views and conclusions contained in this paper are of the authors and should not be interpreted as representing the official policies, either expressed or implied, of the UK MOD or the UK Government.}}

\address{Department of Electronics and Electrical Engineering, University of Liverpool$^{*}$\\
        Department of Computer and Information Science, Link\"oping University$^{\dagger}$}
\author{\name Andrew Millard\email amill212@liverpool.ac.uk \\
      \addr University of Liverpool \\
      \ANDt
      \name Joshua Murphy \email jmurph98@liverpool.ac.uk \\
      \addr University of Liverpool \\
      \ANDt
      \name Zheng Zhao \email zheng.zhao@liu.se\\
      \addr University of Link\"{o}ping \\
      \ANDt
      \name Simon Maskell  \email smaskell@liverpool.ac.uk\\
      \addr University of Liverpool \\
    }
\begin{document}
%
\maketitle
\begin{abstract}
    Partial Bayesian neural networks (pBNNs) have been shown to perform competitively with fully Bayesian neural networks while only having a subset of the parameters be stochastic. Using sequential Monte Carlo (SMC) samplers as the inference method for pBNNs gives a non-parametric probabilistic estimation of the stochastic parameters, and has shown improved performance over parametric methods. We introduce a new SMC-based training method for pBNNs by utilising a guided proposal and incorporating gradient-based Markov kernels, which gives us better scalability on high dimensional problems. We show that our new method outperforms the state-of-the-art (SOTA) in terms of optimal loss. 
\end{abstract}


\section{Introduction}

In recent years, pBNNs have been proposed as a potential ``model solution'' to the high dimensionality issue \cite{sharma2023bayesian}. pBNNs consider a subset of parameters stochastic and, despite this reduced dimensionality, have been shown to outperform fully Bayesian neural networks (BNNs) \cite{sharma2023bayesian, lim2024empiricalimpactneuralparameter}. More formally, let $f(\mathbf{x}; \bm{\theta}, \bm{\psi})$ be a neural network governed by deterministic parameters ($\bm{\psi}$) and stochastic parameters ($\bm{\theta}$) initially generated from a prior $q_0(\bm{\theta})$. If we have a dataset ${(\mathbf{x}_n, \mathbf{y}_n)}_{n=1}^N$ with likelihood $p(\mathbf{y}_n|\bm{\theta},\bm{\psi})$, then training has two objectives: learn the deterministic parameters from the dataset and compute the posterior $p(\bm{\theta} | \mathbf{y}_{1:N}, \bm{\psi})$. Previous work \cite{sharma2023bayesian} shows that placing uncertainty on the first layer of the pBNN empirically yields the best output. This may be because most randomness is aleatoric and originates from the data. Choosing a parametric distribution to model this uncertainty correctly is difficult, motivating non-parametric modeling.

In this paper, we introduce the guided open-horizon sequential Monte Carlo (GOHSMC) algorithm. It builds upon the open-horizon sequential Monte Carlo (OHSMC) sampler \cite{zhao2024feynmankac} by introducing gradient-based Markov kernels and a valid weight update into the pBNN training process to help us more efficiently navigate the high dimensional spaces encountered in neural networks (NNs). We then analyse performance of these SMC methods on benchmark regression datasets. We find our method outperforms the original OHSMC method \cite{zhao2024feynmankac}, with improvements particularly pronounced when the dimensionality of the stochastic layer is high.


\section{Sequential Monte Carlo Samplers} \label{sec:smc_sampler_tutorial}

SMC samplers are Bayesian inference algorithms used when direct sampling from the posterior is difficult. They approximate a target distribution using a weighted set of samples drawn from a sequence of intermediate distributions. 

We aim to sample from a (possibly unnormalised) target distribution $\pi(\bm{\theta})$. Initially, $J$ particles are drawn from a prior and weighted   
\begin{equation}
    \bm{\theta}_0^{(j)} \sim q_0(\cdot), \qquad
    \mathbf{w}_0^{(j)} = \frac{\pi(\bm{\theta}^{(j)}_0)}{q_0(\bm{\theta}^{(j)}_0)}.
    \label{equ:init_sample}
\end{equation}

The algorithm runs for $T$ iterations. At each step $t$, weights are normalised:  
\begin{equation}
    \Tilde{\mathbf{w}}_t^{(j)} = \frac{\mathbf{w}_t^{(j)}}{\sum_{j=1}^J \mathbf{w}_t^{(j)}}. 
    \label{equ:normalise}
\end{equation}

To avoid particle degeneracy, resampling is triggered when the effective sample size  
\begin{equation}
    J_\mathrm{eff} = \frac{1}{\sum^{J}_{i=1} (\Tilde{\mathbf{w}}_t^{(j)})^2}, 
    \label{equ:jeff}
\end{equation}
falls below $J/2$. We use multinomial resampling \cite{inproceedings}, after which weights are reset to $1/J$. New particles are proposed using a Markov kernel:  
\begin{equation}
    \bm{\theta}_t^{(j)} \sim q_t^\theta(\cdot|\bm{\theta}^{(j)}_{t-1}),
\end{equation}
which propagates particles from iteration $t-1$ to $t$. Weights are then updated via  
\begin{equation}
    \mathbf{w}^{(j)}_t = \mathbf{w}^{(j)}_{t-1} 
    \frac{\pi(\bm{\theta}^{(j)}_t)}{\pi(\bm{\theta}^{(j)}_{t-1})}
    \frac{L_t^\theta(\bm{\theta}^{(j)}_{t-1}| \bm{\theta}^{(j)}_{t})}{q_t^\theta(\bm{\theta}^{(j)}_{t}| \bm{\theta}^{(j)}_{t-1})},
    \label{equ:fwd-bwd-kern}
\end{equation}
where $L_t^\theta(\bm{\theta}_{t-1}|\bm{\theta}_{t})$ is the backward (L) kernel. If resampling occurred at step $t$, the previous weight in~\eqref{equ:fwd-bwd-kern} is $1/J$. 

For a detailed overview of SMC methods, see \cite{del2006sequential}. In this paper we use the term weights to mean the weights of samples in an SMC sampler (Section \ref{sec:smc_sampler_tutorial}), not neural network parameters. In this paper, parameters refer to values associated with neural network nodes.

The terms particles and samples are used interchangeably in the SMC literature. We adopt both, choosing the more common term for each context (e.g. particle degeneracy rather than sample degeneracy).

\subsection{SMC for pBNNs, Stochastic Gradient and Open-horizon SMC}
SMC samplers can be applied to sample the posterior distribution of the stochastic part of the pBNN with $\pi(\bm{\theta}, \bm{\psi})=  p(\mathbf{y}_n|\bm{\theta},\bm{\psi}) q_0(\bm{\theta}) \propto p(\bm{\theta} \mid \mathbf{y}_{1:N}, \bm{\psi})$. However, it remains to be shown how to learn the deterministic part of the pBNN, in particular, when there is a large number of data-points $D$. The \emph{SMC sampler for pBNN} (SMC-pBNN) algorithm given in \cite{zhao2024feynmankac} loops over the entire dataset and computes gradients after a full pass of the dataset. this guarantees accurate gradient estimation and posterior convergence, but incurs high computational cost.

The Stochastic Gradient SMC (SGSMC) Algorithm builds upon this by instead using a mini-batch of data. If we denote a batch size $M$ where $1 \leq M \leq D$ then let $\mathbf{S}_M := \{S_M(1), S_M(2), \ldots , S_M(M)\}$ be a set of batch indices. We then approximate our log-likelihood with respect to $\bm{\psi}$ as
\begin{equation}
    \log p(\mathbf{y}_{1:N}| \bm{\psi}) \approx \frac{N}{M} \log p(\mathbf{y}_{\mathbf{S}_M}| \bm{\psi}),
\end{equation}
by only considering the subset of data $\mathbf{y}_{\mathbf{S}_M}$. Moreover, the SMC sampler also allows us to simultaneously compute an approximation of the gradient of the log-likelihood 
\begin{align}
    \frac{N}{M}\nabla_{\bm\psi} \log p(&\mathbf{y}_{\mathbf{S}_M}| \bm{\psi}) \nonumber\\
    &= \frac{N}{M} \mathbb{E}_{p(\bm{\theta}|\mathbf{y}_{\mathbf{S}_M},\bm{\psi})} \left[ \nabla_{\bm\psi} \log p(\mathbf{y}_{\mathbf{S}_M}| \bm{\theta}, \bm{\psi}) \right] \nonumber\\
    &\approx \frac{N}{M} \sum_{j=1}^J\mathbf{\tilde{w}}^{(j)} \nabla \log p(\mathbf{y}_\mathbf{{S}_M} | \bm{\theta}^{(j)}, \bm{\psi}).
    \label{equ:approx-grad}
\end{align}
Note that this approximation is biased due to the presence of the normalised weights, $\mathbf{\tilde{w}}^{(j)}$. 

This stochastic gradient is computed using the subdataset $\mathbf{y}_{\mathbf{S}_M}$, and the expectation is taken over the random batch indices.  The SGSMC algorithm updates the particles and weights using the stochastic gradient approximation.  At each iteration of the gradient optimisation, a subdataset is sampled then the SMC sampler is applied on this subdataset to estimate the gradient and update the pBNN stochastic parameters.  However, the SGSMC algorithm does not really sample the posterior $p(\bm{\theta} \mid \mathbf{y}_{1:N}, \bm{\psi})$ which motivated the OHSMC sampler introduced in \cite{zhao2024feynmankac}. 

OHSMC merges the loop of the stochastic optimisation for $\bm{\psi}$ and the SMC sampling loop in a principled way.  Specifically, at each iteration of OHSMC, the algorithm randomly samples a subdataset, and then approximates the gradient~\eqref{equ:approx-grad} and the target posterior distribution concurrently. Crucially, the OHSMC sampler can process the subdataset in parallel, while SGSMC sequentially loops over the elements in the subdataset. Empirically, OHSMC also provides a better approximation to the target posterior distribution by linking the posterior distribution estimates across iterations. Unlike SGSMC, which independently restarts from the prior distribution at each step, OHSMC uses the posterior from the previous iteration as the starting point for the next. This warm-start strategy improves computational efficiency and maintains continuity in the posterior distribution estimation.

\subsection{Guided OHSMC}
\begin{algorithm}[tb]
    \caption{GOHSMC}\label{alg:GOHSMC_example}
\begin{algorithmic}
    \REQUIRE Training data ${(\mathbf{x}_n, \mathbf{y}_n)}_{n=1}^N$, number of samples $J$, initial parameters $\bm{\psi}_0$, prior distribution $q_0(\cdot)$, learning rate $\epsilon$, batch size $M$
    \ENSURE The MLE estimate $\bm{\psi}_t$
    \STATE Initialize samples $\{\bm{\theta}_{0}^{(j)}\}_{j=1}^J \sim q_0(\cdot)$
    \STATE Initialize weights according to \eqref{equ:pbnn_init}
    \FOR {$t = 1, 2, \ldots$ until convergence}
        \STATE Draw sub dataset $\mathbf{y}_{\mathbf{S}_M(t)} \subset \mathbf{y}_{1:N}$
        \STATE Calculate $J_\mathrm{eff}$ using \eqref{equ:jeff}
        \IF{$J_\mathrm{eff} < J/2$}
            \STATE Resample $[\bm{\theta}^{(1)}_t ... \bm{\theta}^{(J)}_t]$ with probability $[\Tilde{\mathbf{w}}^{(1)}_t ... \Tilde{\mathbf{w}}^{(J)}_t]$
            \STATE Reset all weights to $\frac{1}{J}$
        \ENDIF
        \FOR {$j=1$ {\bfseries to} $J$}
            \STATE Propagate particles $\bm{\theta}_{t}^{(j)} \sim q_t^\theta(\cdot|\bm{\theta}_{t-1}^{(j)}, \bm{\psi}_{t-1})$
            \STATE Update weight $\mathbf{w}_{t}^{(j)}$ with \eqref{equ:langevin_pbnn_w_eq}
        \ENDFOR
        \STATE Normalize weights using (3)
        \STATE $g(\bm{\psi}_{t-1}) = \frac{N}{M} \sum_{j=1}^J \mathbf{\tilde{w}}_{t}^{(j)} \nabla \log p(\mathbf{y}_\mathbf{{S}_t^M} | \bm{\theta}_{t}^{(j)}, \bm{\psi}_{t-1})$
        \STATE Update parameter $\bm{\psi}_t = \bm{\psi}_{t-1} + \epsilon g(\bm{\psi}_{t-1})$
    \ENDFOR
\end{algorithmic}
\end{algorithm}
The main criticism of the original OHSMC sampler is that it is based on a bootstrap version of an SMC sampler. Namely, they invoke a Markov kernel that leaves invariant the \emph{previous} posterior distribution. We hereby build upon OHSMC to develop GOHSMC in Algorithm~\ref{alg:GOHSMC_example} by making the Markov kernel leave invariant the \emph{current} posterior distribution. This gives us a guided improvement of the original OHSMC by incorporating information from the target posterior distribution. Consequently, this guided version yields a more effective importance proposal leading to better statistical efficiency, as evidenced in the new weight update ~\eqref{equ:langevin_pbnn_w_eq}. Another improvement we deliver is better scalability in high-dimensional $\bm{\theta}$. Unlike OHSMC which essentially uses a Random Walk (RW) Markov kernel \cite{metropolis1953equation} \cite{givens1996local}, we use gradient-based Markov kernels as proposals, in particular, the unadjusted Langevin dynamics (LD) to better explore the high-dimensional latent space \cite{neal2012bayesian, liu2001monte}, 
\begin{align}
  \mathbf{P}_{t-\frac{1}{2}} &= \mathbf{P}_{t-1} + \frac{\epsilon}{2} \nabla \log \pi(\bm{\theta}_{t-1}), \label{eq:mom_update}\\
  \bm{\theta}_{t} &= \bm{\theta}_{t-1} + \epsilon \mathbf{P}_{t-\frac{1}{2}}, \label{eq:pos_step} \\
  \mathbf{P}^* &= \mathbf{P}_{t-\frac{1}{2}} + \frac{\epsilon}{2} \nabla \log \pi(\bm{\theta}_{t}), \label{eq:full_step} 
\end{align}
\noindent where $\mathbf{P}^{(j)}_{t-1}$ is the initial momentum drawn from a Gaussian $\mathcal{N}(0,\bm{M})$ and $\bm{M}$ is known as a mass matrix but often set to Identity ($\bm{M} = \bm{I}$). Technically, only \eqref{eq:mom_update} and \eqref{eq:pos_step} are required for unadjusted LD. However, completing the momentum update facilitates a cancellation in the determinants when using the so-called \emph{forwards-proposal} (FP) L-kernel \cite{murphy2025hessmc2sequentialmontecarlo} thanks to the reversibility of the LD process \cite{Dai2022}. This gives us the tractable weight update
\begin{equation}
    \mathbf{w}^{(j)}_t = \mathbf{w}^{(j)}_{t-1} \frac{\pi(\bm{\theta}^{(j)}_t)}{\pi(\bm{\theta}^{(j)}_{t-1})}\frac{L_t^P(-\mathbf{P}^{(j)*})}{q_t^P(\mathbf{P}^{(j)}_{t-1})}.
    \label{equ:langevin_w_eq}
\end{equation}

Our weight update is also conditional on the deterministic parameters, so we alter it for the pBNN context. The gradient of the LD is conditioned on the deterministic parameters which are the same for both the forwards and backwards moves. Therefore the LD in this scenario are described by
\begin{align}
  \mathbf{P}_{t-\frac{1}{2}} &= \mathbf{P}_{t-1} + \frac{\epsilon}{2} \nabla \log \pi(\bm{\theta}_{t-1}|\bm{\psi}_{t-1}), \\
  \bm{\theta}_{t} &= \bm{\theta}_{t-1} + \epsilon \mathbf{P}_{t-\frac{1}{2}}, \\
  \mathbf{P}^* &= \mathbf{P}_{t-\frac{1}{2}} + \frac{\epsilon}{2} \nabla \log \pi(\bm{\theta}_{t}|\bm{\psi}_{t-1}),  
\end{align}
which can also be viewed as the Hamiltonian Monte Carlo (HMC) dynamics with a trajectory length of $S=1$. 

Leapfrog can be considered a function which uses the position $\bm{\theta}_{t-1}$ and momentum $\mathbf{P}_{t-1}$ to transform the state i.e. ${\bm{\theta}_{t} = f_{LMC}(\bm{\theta}_{t-1}, \mathbf{P}_{t-1})}$. We define our proposal in the context of a pBNN as
\begin{align}
    q_t^\theta(&f_{LMC}(\bm{\theta}_{t-1}, \mathbf{P}_{t-1})|\bm{\theta}_{t-1}, \bm{\psi}_{t-1}). \label{eq:lv_before_change}
\end{align}
We use a change of variables described by 
\begin{align}
    Y &= g(X), \\
    p_Y(y) &= p_X(x) \bigg| \frac{d g(X)}{d X} \bigg|^{-1},
    \label{eq:change_of_variables}
\end{align}
where $Y = \bm{\theta}_t$, $X=\mathbf{P}_{t-1}$ and $g(X) = f_{LMC}(\bm{\theta}_{t-1}, \mathbf{P}_{t-1})$ to re-write the proposal as
\begin{align}
    q_t^\theta(f_{LMC}(&\bm{\theta}_{t-1}, \mathbf{P}_{t-1})|\bm{\theta}_{t-1}, \bm{\psi}_{t-1}) \nonumber\\
    &= q_t^P(\mathbf{P}_{t-1}) \begin{vmatrix} 
            \frac{df_{LMC}(\bm{\theta}_{t-1}, \mathbf{P}_{t-1})}{d\mathbf{P}_{t-1}} \end{vmatrix}^{-1}, \nonumber\\
        &= \mathcal{N}(\mathbf{P}_{t-1}; 0,\bm{M}) \begin{vmatrix} 
            \frac{df_{LMC}(\bm{\theta}_{t-1}, \mathbf{P}_{t-1})}{d\mathbf{P}_{t-1}} \end{vmatrix}^{-1}.\label{eq:lv_after_change}
\end{align} 
as momentum is drawn from a Gaussian distribution independent of both $\bm{\theta}$ and $\bm{\psi}$, thus
%
For the L-kernel we have
\begin{align}
        L_t^\theta(f_{LMC}(&\bm{\theta}_t, -\mathbf{P}^*) |\bm{\theta}_ t, \bm{\psi}_{t-1}) \nonumber\\
        &= \mathcal{N}(-\mathbf{P}^*; 0,\bm{M}) \begin{vmatrix}
            \frac{df_{LMC}(\bm{\theta}_t, -\mathbf{P}^*)}{d\mathbf{P}^*}
        \end{vmatrix}^{-1}, \label{eq: l=q}
\end{align}
leaving us with the final weight update: 
\begin{equation}
    \mathbf{w}^{(j)}_t = \mathbf{w}^{(j)}_{t-1} \frac{\pi(\bm{\theta}^{(j)}_{t} | \bm{\psi}_{t-1})}{\pi(\bm{\theta}^{(j)}_{t-1} | \bm{\psi}_{t-2})} \frac{\mathcal{N}(-\mathbf{P}^*; 0,\bm{M})}{\mathcal{N}(\mathbf{P}_{t-1}; 0,\bm{M})},
    \label{equ:langevin_pbnn_w_eq}
\end{equation} 
and 
\begin{equation}
    \mathbf{w}^{(j)}_0 = \frac{\pi(\bm{\theta}^{(j)}_{0} | \bm{\psi}_{0})}{q_0^\theta( \bm{\theta}^{(j)}_{0})}.
    \label{equ:pbnn_init}
\end{equation} 
Note that there is no optimisation of the initial deterministic parameters, $\bm{\psi}_{0}$. Further details of this weight update can be found in \cite{devlin2024NUTS}. As unadjusted LD do not leave the target distribution invariant, we use importance weights to correct for this and ensure the correct target distribution is recovered.

Let $C_f$ be the cost of a single forward pass through the NN, which is used to calculate the likelihoods for the weights, and $C_b$ the cost of a backward pass to compute the gradient of the loss, as is used for the LD proposal. For OHSMC, the cost per iteration is $\mathcal{O}\!\left(J\,C_f\right)$ but for GOHSMC, each move requires a forward and backward pass giving $\mathcal{O}\!\left(2J\,(C_f + C_b)\right)$.


\section{Experiments}

\begin{table*}[t!]
\tiny
\centering
\resizebox{0.95\textwidth}{!}{%
\begin{tabular}{lcccccc}
\toprule
\textbf{Method} & \textbf{California} (100) & \textbf{Concrete} (50) & \textbf{Yacht} (50) & \textbf{Red Wine} (50) & \textbf{White Wine} (50) & \textbf{Naval} (50) \\
\midrule
\multicolumn{7}{c}{\textbf{RMSE ($\downarrow$)}} \\
\cmidrule(lr){1-7}
GOHSMC LD & 0.5299 $\pm$ 0.0842 & \textbf{0.3318 $\pm$ 0.1477} & \textbf{0.0766 $\pm$ 0.0790} & \textbf{0.6400 $\pm$ 0.2066} & \textbf{0.7040 $\pm$ 0.2659} & \textbf{0.0017 $\pm$ 0.0010} \\
OHSMC RW & 0.5790 $\pm$ 0.1268 & 0.4199 $\pm$ 0.1866 & 0.1621 $\pm$ 0.1062 & 0.6586 $\pm$ 0.1963 & 0.7204 $\pm$ 0.2685 & 0.0056 $\pm$ 0.0017 \\
SGHMC & 0.5536 $\pm$ 0.1219 & 0.7084 $\pm$ 0.3364 & 0.8658 $\pm$ 0.3992 & 0.6876 $\pm$ 0.2474 & 0.7171 $\pm$ 0.2601 & 0.0159 $\pm$ 0.0083 \\
SWAG & \textbf{0.5232 $\pm$ 0.1059} & 0.3411 $\pm$ 0.1198 & 0.0976 $\pm$ 0.1064 & 0.7101 $\pm$ 0.2443 & 0.7477 $\pm$ 0.3088 & 0.0166 $\pm$ 0.0130 \\
VI & 0.6452 $\pm$ 0.1094 & 0.4124 $\pm$ 0.1867 & 0.2818 $\pm$ 0.3577 & 0.6459 $\pm$ 0.1252 & 0.7442 $\pm$ 0.2702 & 0.0221 $\pm$ 0.0154 \\
SVGD & 0.6373 $\pm$ 0.1266 & 0.6255 $\pm$ 0.2621 & 0.6659 $\pm$ 0.3447 & 0.6619 $\pm$ 0.1639 & 0.7180 $\pm$ 0.2645 & 0.0096 $\pm$ 0.0039 \\
\midrule
\multicolumn{7}{c}{\textbf{R$^2$ / Bias ($\uparrow / \downarrow$)}} \\
\cmidrule(lr){1-7}
GOHSMC LD & 0.7856 / \textbf{0.0036} & 0.8278 / 0.0992 & \textbf{0.9933} / \textbf{0.0054} & 0.2803 / 0.1060 & \textbf{0.3488 / 0.0407} & \textbf{0.9375 / 0.0224} \\
OHSMC RW & 0.7449 / 0.0082 & 0.7525 / 0.0862 & 0.9642 / 0.0187 & 0.2489 / 0.0726 & 0.3197 / 0.0411 & 0.4336 / 0.0488 \\
SGHMC & 0.7707 / 0.0082 & 0.4297 / 0.0425 & 0.0189 / 0.1194 & 0.1936 / 0.0665 & 0.3457 / 0.0391 & -3.5389 / 1.2464 \\
SWAG & \textbf{0.7949} / 0.0066 & \textbf{0.8874 / 0.0185} & 0.9898 / 0.0101 & 0.1769 / 0.1351 & 0.2926 / 0.0671 & -3.8915 / 3.0226 \\
VI & 0.7012 / 0.0393 & 0.6619 / 0.2967 & 0.9129 / 0.1340 & \textbf{0.3750} / 0.3000 & 0.2506 / 0.0795 & -7.5321 / 4.4937 \\
SVGD & 0.6964 / 0.0071 & 0.5214 / 0.0463 & 0.4231 / 0.0873 & 0.2595 / \textbf{0.0459} & 0.3434 / 0.0512 & -0.6495 / 0.2734 \\
\midrule
\multicolumn{7}{c}{\textbf{NLL ($\downarrow$)}} \\
\cmidrule(lr){1-7}
GOHSMC LD & 1.0594 $\pm$ 0.0035 & \textbf{0.9740 $\pm$ 0.0109} & \textbf{0.9219 $\pm$ 0.0031} & 1.1238 $\pm$ 0.0213 & 1.1668 $\pm$ 0.0353 & \textbf{0.9189 $\pm$ 0.0000}\\
OHSMC RW & 1.0866 $\pm$ 0.0080 & 1.0071 $\pm$ 0.0174 & 0.9321 $\pm$ 0.0056 & 1.1358 $\pm$ 0.0193 & 1.1784 $\pm$ 0.0361 & 0.9190 $\pm$ 0.0000\\
SGHMC & 1.0720 $\pm$ 0.0074 & 1.2159 $\pm$ 0.0326 & 1.2938 $\pm$ 0.0797 & 1.1679 $\pm$ 0.0254 & 1.1766 $\pm$ 0.0333 & 0.9191 $\pm$ 0.0000\\
SWAG & \textbf{1.0558 $\pm$ 0.0056} & 0.9771 $\pm$ 0.0072 & 0.9237 $\pm$ 0.0057 & 1.1711 $\pm$ 0.0298 & 1.1984 $\pm$ 0.0477 & 0.9191 $\pm$ 0.0001\\
VI & 1.1273 $\pm$ 0.0240 & 0.9778 $\pm$ 0.0536 & 0.9586 $\pm$ 0.0640 & \textbf{1.0830 $\pm$ 0.0444} & \textbf{1.1645 $\pm$ 0.0680} & 0.9192 $\pm$ 0.0001\\
SVGD & 1.1215 $\pm$ 0.0077 & 1.1694 $\pm$ 0.0387 & 1.1406 $\pm$ 0.0594 & 1.1471 $\pm$ 0.0126 & 1.1774 $\pm$ 0.0347 & 0.9190 $\pm$ 0.0000\\
\midrule
\multicolumn{7}{c}{\textbf{CRPS ($\downarrow$)}} \\
\cmidrule(lr){1-7}
GOHSMC LD & 0.3313 $\pm$ 0.0026 & \textbf{0.2762 $\pm$ 0.0083} & \textbf{0.2360 $\pm$ 0.0025} & 0.3793 $\pm$ 0.0137 & 0.4089 $\pm$ 0.0215 & \textbf{0.2337 $\pm$ 0.0000}\\
OHSMC RW & 0.3491 $\pm$ 0.0052 & 0.3005 $\pm$ 0.0131 & 0.2440 $\pm$ 0.0044 & 0.3878 $\pm$ 0.0120 & 0.4154 $\pm$ 0.0218 & 0.2337 $\pm$ 0.0000\\
SGHMC & 0.3398 $\pm$ 0.0052 & 0.4452 $\pm$ 0.0236 & 0.4776 $\pm$ 0.0450 & 0.4080 $\pm$ 0.0166 & 0.4145 $\pm$ 0.0210 & 0.2338 $\pm$ 0.0000\\
SWAG & \textbf{0.3284 $\pm$ 0.0041} & 0.2776 $\pm$ 0.0049 & 0.2374 $\pm$ 0.0044 & 0.4042 $\pm$ 0.0160 & 0.4263 $\pm$ 0.0273 & 0.2338 $\pm$ 0.0001\\
VI & 0.3731 $\pm$ 0.0126 & 0.2787 $\pm$ 0.0398 & 0.2595 $\pm$ 0.0394 & \textbf{0.3537 $\pm$ 0.0330} & \textbf{0.4048 $\pm$ 0.0425} & 0.2339 $\pm$ 0.0001\\
SVGD & 0.3709 $\pm$ 0.0053 & 0.4127 $\pm$ 0.0229 & 0.3887 $\pm$ 0.0379 & 0.3944 $\pm$ 0.0078 & 0.4157 $\pm$ 0.0220 & 0.2337 $\pm$ 0.0000\\
\bottomrule
\end{tabular}
}
\caption{Predictive accuracy and uncertainty quantification metrics across datasets. Batch size shown in brackets.}
\label{tab:all_metrics_combined}
\end{table*}

Experiments were conducted using JAX on an NVIDIA A100 GPU. Stochastic parameters were initialised from a standard normal $\mathcal{N}(0, \mathbf{I})$; deterministic parameters used JAX’s default initialisation. In tables, our method is labelled GOHSMC LD. We evaluated on 6 UCI regression datasets: Red Wine Quality, White Wine Quality, California Housing, Concrete Compressive Strength, Yacht Hydrodynamics, and Naval Propulsion. Baselines were OHSMC RW, Variational Inference (VI) \cite{graves2011practical}, Stochastic Gradient HMC (SGHMC) \cite{chen2014stochastic}, Stochastic Weight Averaging Gaussian (SWAG) \cite{maddox2019simplebaselinebayesianuncertainty}, and Stein Variational Gradient Descent (SVGD) \cite{NIPS2017_17ed8abe}. The models were feed-forward networks with three dense layers and GeLU activations. First-layer widths were 350 (Yacht), 450 (California, Concrete), 600 (Red/White Wine), and 900 (Naval) with the first layer sampled and remaining layers optimised with Adam \cite{kingma2015adam} (learning rate 0.01). RW and LD kernel scales were 0.01 and $1/N_{data}$ respectively; all methods used 100 samples. Each run was 100 epochs, with the best validation-loss parameters evaluated on the test set. Data was split 60\%/30\%/10\% for train/validation/test, and results averaged over 5 runs. We report root mean square error (RMSE), R$^2$, bias, negative log-likelihood (NLL) and continuous ranked probability score (CRPS) with standard deviations in Table~\ref{tab:all_metrics_combined}.

GOHSMC LD had the best RMSE on five of the six experiments and was close to the top-performing method, SWAG, on the California dataset. The benefits are less pronounced on other metrics but GOHSMC LD is still the best performing method on at least three experiments. Note that GOHSMC outperforms OHSMC in nearly every case.  While some overlap in error bars exists, the consistent performance of GOHSMC across all datasets indicates potential overall benefits. 

\section{Conclusions and Further Work}
We have introduced gradient based proposals into SMC samplers to train pBNNs. Specifically we have introduced LD as the Markov kernel with a FP L-kernel and demonstrated on six benchmark datasets that we can outperform the current SOTA SMC methods. Although GOHSMC provides the best performance overall, there are overlaps in the standard deviations for certain datasets. Additional Monte Carlo runs may further reduce variance, strengthening the evidence that GOHSMC definitively has the best performance.

GOHSMC provides a simple yet flexible framework and as such is well suited for future research where more sophisticated proposals can be explored for sample propagation. So far, we have only introduced LD as a gradient based proposal. However, with an FP L-kernel, there are other gradient based proposals that are also worth exploring such as HMC \cite{neal2011mcmc}. One issue with HMC is the need to tune the hyperparameters. Different approaches to solving this could be taken; the No U-Turn algorithm could be used as a Markov kernel \cite{devlin2024NUTS}. This algorithm has a U-turn termination criteria which adaptively tunes the number of leapfrog steps. Alternatively, using the ChEES criterion \cite{pmlr-v130-hoffman21a} in an SMC sampler \cite{millard2025incorporatingcheescriterionsequential} would allow us to tune the trajectory length during warm-up. 

\newpage

\bibliographystyle{IEEEbib}
\bibliography{main}

@article{del2006sequential,
  title = {{Sequential {Monte Carlo} Samplers}},
  author = {Del Moral, Pierre and Doucet, Arnaud and Jasra, Ajay},
  journal = {Journal of the Royal Statistical Society Series B: Statistical Methodology},
  volume = {68},
  number = {3},
  pages = {411--436},
  year = {2006},
  publisher = {Oxford University Press}
}

@inbook{neal2011mcmc,
  title = {{{MCMC} Using {Hamiltonian} Dynamics}},
  author = {Neal, Radford M.},
  booktitle = {Handbook of {Markov Chain Monte Carlo}},
  pages = {113--162},
  year = {2011},
  publisher = {Chapman and Hall/CRC},
  doi = {10.1201/b10905}
}

@article{kingma2015adam,
  title={Adam: A method for stochastic optimization},
  author={Kingma, Diederik P},
  journal={arXiv preprint arXiv:1412.6980},
  year={2014}
}

@inproceedings{maddox2019simplebaselinebayesianuncertainty,
 author = {Maddox, Wesley J and Izmailov, Pavel and Garipov, Timur and Vetrov, Dmitry P and Wilson, Andrew Gordon},
 booktitle = {Advances in Neural Information Processing Systems},
 editor = {H. Wallach and H. Larochelle and A. Beygelzimer and F. d\textquotesingle Alch\'{e}-Buc and E. Fox and R. Garnett},
 pages = {},
 publisher = {Curran Associates, Inc.},
 title = {{A Simple Baseline for Bayesian Uncertainty in Deep Learning}},
 url = {https://proceedings.neurips.cc/paper_files/paper/2019/file/118921efba23fc329e6560b27861f0c2-Paper.pdf},
 volume = {32},
 year = {2019}
}

@InProceedings{sharma2023bayesian,
  title = 	 {{Do Bayesian Neural Networks Need To Be Fully Stochastic?}},
  author =       {Sharma, Mrinank and Farquhar, Sebastian and Nalisnick, Eric and Rainforth, Tom},
  booktitle = 	 {Proceedings of The 26th International Conference on Artificial Intelligence and Statistics},
  pages = 	 {7694--7722},
  year = 	 {2023},
  editor = 	 {Ruiz, Francisco and Dy, Jennifer and van de Meent, Jan-Willem},
  volume = 	 {206},
  series = 	 {Proceedings of Machine Learning Research},
  month = 	 {25--27 Apr},
  publisher =    {PMLR},
  url = 	 {https://proceedings.mlr.press/v206/sharma23a.html}
}

@InProceedings{zhao2024feynmankac,
  title = 	 {{On {F}eynman-{K}ac Training of Partial {B}ayesian Neural Networks}},
  author =       {Zhao, Zheng and Mair, Sebastian and B. Sch\"{o}n, Thomas and Sj\"{o}lund, Jens},
  booktitle = 	 {Proceedings of The 27th International Conference on Artificial Intelligence and Statistics},
  pages = 	 {3223--3231},
  year = 	 {2024},
  editor = 	 {Dasgupta, Sanjoy and Mandt, Stephan and Li, Yingzhen},
  volume = 	 {238},
  series = 	 {Proceedings of Machine Learning Research},
  month = 	 {02--04 May},
  publisher =    {PMLR},
  url = 	 {https://proceedings.mlr.press/v238/zhao24b.html}
}

@article{metropolis1953equation,
  title = {{Equation of State Calculations by Fast Computing Machines}},
  author={Metropolis, N. and Rosenbluth, A. W. and Rosenbluth, M. N. and Teller, A. H. and Teller, E.},
  journal={Journal of Chemical Physics},
  volume={21},
  number={6},
  pages={1087--1092},
  year={1953},
  doi={10.1063/1.1699114}
}

@article{givens1996local,
  title = {{Local Adaptive Importance Sampling for Multivariate Densities with Strong Nonlinear Relationships}},
  author={Givens, Geof H and Raftery, Adrian E},
  journal={Journal of the American Statistical Association},
  volume={91},
  number={433},
  pages={132--141},
  year={1996},
  publisher={Taylor \& Francis}
}

@article{Dai2022,
  title = {{An Invitation to Sequential {Monte Carlo} Samplers}},
  author = {Chenguang Dai and Jeremy Heng and Pierre E. Jacob and Nick Whiteley},
  journal = {Journal of the American Statistical Association},
  volume = {117},
  number = {539},
  pages = {1587--1600},
  year = {2022}
}

@inproceedings{chen2014stochastic,
  title = {{Stochastic Gradient Hamiltonian Monte Carlo}},
  author = {Chen, Tianqi and Fox, Emily and Guestrin, Carlos},
  booktitle = {International Conference on Machine Learning (ICML)},
  pages = {1683--1691},
  year = {2014},
  organization = {PMLR}
}

@inproceedings{inproceedings,
  title={Comparison of resampling schemes for particle filtering},
  author={Douc, Randal and Capp{\'e}, Olivier},
  booktitle={2005 International Symposium on Image and Signal Processing and Analysis, (ISPA).},
  pages={64--69},
  year={2005},
  organization={IEEE}
}

@book{neal2012bayesian,
  title = {{Bayesian Learning for Neural Networks}},
  author = {Neal, Radford M.},
  volume = {118},
  year = {2012},
  publisher = {Springer Science \& Business Media}
}

@book{liu2001monte,
  title = {{Monte Carlo Strategies in Scientific Computing}},
  author = {Liu, Jun S.},
  volume = {10},
  year = {2001},
  publisher = {Springer}
}

@ARTICLE{devlin2024NUTS,
  title = {{The {No-U-Turn Sampler} as a Proposal Distribution in a Sequential {Monte Carlo} Sampler Without Accept/Reject}},
  author = {Devlin, Lee and Carter, Matthew and Horridge, Paul and Green, Peter L. and Maskell, Simon},
  journal = {IEEE Signal Processing Letters},
  year = {2024},
  volume = {31},
  number = {},
  pages = {1089--1093},
  doi = {10.1109/LSP.2024.3386494}
}

@inproceedings{graves2011practical,
 author = {Graves, Alex},
 booktitle = {Advances in Neural Information Processing Systems},
 editor = {J. Shawe-Taylor and R. Zemel and P. Bartlett and F. Pereira and K.Q. Weinberger},
 pages = {},
 publisher = {Curran Associates, Inc.},
 title = {{Practical Variational Inference for Neural Networks}},
 url = {https://proceedings.neurips.cc/paper_files/paper/2011/file/7eb3c8be3d411e8ebfab08eba5f49632-Paper.pdf},
 volume = {24},
 year = {2011}
}

@article{lim2024empiricalimpactneuralparameter,
      title={{The Empirical Impact of Neural Parameter Symmetries, or Lack Thereof}}, 
      author={Derek Lim and Theo Moe Putterman and Robin Walters and Haggai Maron and Stefanie Jegelka},
      year={2024},
      eprint={2405.20231},
      journal = {arXiv preprint arXiv:2405.20231},
      archivePrefix={arXiv},
      primaryClass={cs.LG},
      url={https://arxiv.org/abs/2405.20231}, 
}

@inproceedings{NIPS2017_17ed8abe,
 author = {Liu, Qiang},
 booktitle = {Advances in Neural Information Processing Systems},
 editor = {I. Guyon and U. Von Luxburg and S. Bengio and H. Wallach and R. Fergus and S. Vishwanathan and R. Garnett},
 pages = {},
 publisher = {Curran Associates, Inc.},
 title = {{Stein Variational Gradient Descent as Gradient Flow}},
 url = {https://proceedings.neurips.cc/paper_files/paper/2017/file/17ed8abedc255908be746d245e50263a-Paper.pdf},
 volume = {30},
 year = {2017}
}

@InProceedings{pmlr-v130-hoffman21a,
  title = 	 {{An Adaptive-MCMC Scheme for Setting Trajectory Lengths in Hamiltonian Monte Carlo}},
  author =       {Hoffman, Matthew and Radul, Alexey and Sountsov, Pavel},
  booktitle = 	 {Proceedings of The 24th International Conference on Artificial Intelligence and Statistics},
  pages = 	 {3907--3915},
  year = 	 {2021},
  editor = 	 {Banerjee, Arindam and Fukumizu, Kenji},
  volume = 	 {130},
  series = 	 {Proceedings of Machine Learning Research},
  month = 	 {13--15 Apr},
  publisher =    {PMLR},
  url = 	 {https://proceedings.mlr.press/v130/hoffman21a.html}
}

@inproceedings{millard2025incorporatingcheescriterionsequential,
  author={Millard, Andrew and Murphy, Joshua and Frisch, Daniel and Maskell, Simon},
  booktitle={2025 28th International Conference on Information Fusion (FUSION)}, 
  title={{Incorporating the ChEES Criterion Into Sequential Monte Carlo Samplers}}, 
  year={2025},
  volume={},
  number={},
  pages={1-8},
  doi={10.23919/FUSION65864.2025.11124101}}

@inproceedings{murphy2025hessmc2sequentialmontecarlo,
  title={{Hess-MC$^2$: Sequential Monte Carlo Squared Using Hessian Information and Second Order Proposals}},
  author={Murphy, Joshua and Rosato, Conor and Millard, Andrew and Devlin, Lee and Horridge, Paul and Maskell, Simon},
  booktitle={2025 IEEE 35th International Workshop on Machine Learning for Signal Processing (MLSP)},
  pages={1--6},
  year={2025},
  organization={IEEE}
}

\end{document}